\documentclass[conference]{IEEEtran}
\IEEEoverridecommandlockouts
\usepackage{amsmath,amssymb,amsfonts}
\usepackage{algorithmic}
\usepackage{graphicx}
\usepackage{textcomp}
\usepackage{xcolor}
\usepackage{times}  % DO NOT CHANGE THIS
\usepackage{helvet}  % DO NOT CHANGE THIS
\usepackage{courier}  % DO NOT CHANGE THIS
\usepackage[T1]{fontenc}
\usepackage[utf8]{inputenc}
\usepackage[hyphens]{url}  % DO NOT CHANGE THIS
\usepackage{graphicx} % DO NOT CHANGE THIS
\usepackage{natbib}  % DO NOT CHANGE THIS AND DO NOT ADD ANY OPTIONS TO IT
\usepackage{caption} % DO NOT CHANGE THIS AND DO NOT ADD ANY OPTIONS TO IT
\usepackage{amsmath,amssymb,amsfonts, amsthm}
\usepackage{algorithmic}
\usepackage{algorithm}
\usepackage{comment}
\usepackage{booktabs}       % professional-quality tables
\usepackage{nicefrac}       % compact symbols for 1/2, etc.
\usepackage{microtype}      % microtypography
\usepackage{xcolor}         % colors
\usepackage{etoolbox}
\usepackage{changepage}
\usepackage{float}
\usepackage{ragged2e}
\usepackage{subfigure}
\usepackage{verbatim}
\usepackage{diagbox}
\usepackage{threeparttable}
\usepackage{multirow}
\usepackage{multicol}

\theoremstyle{plain}
\newtheorem{theorem}{Theorem}
\newtheorem{proposition}[theorem]{Proposition}
\newtheorem{lemma}[theorem]{Lemma}

\theoremstyle{definition}
\newtheorem{definition}[theorem]{Definition}

\theoremstyle{remark}
\newtheorem{remark}[theorem]{Remark}

\def\BibTeX{{\rm B\kern-.05em{\sc i\kern-.025em b}\kern-.08em
    T\kern-.1667em\lower.7ex\hbox{E}\kern-.125emX}}

\begin{document}

\title{DeepDefense: Robust Learning via Layer-Wise Gradient-Feature Alignment}

\author{\IEEEauthorblockN{Ci Lin, Tet Yeap, Iluju Kiringa}
	\IEEEauthorblockA{\textit{Electrical Engineering and Computer Science} \\
		\textit{University of Ottawa}\\
		\{clin072, tyeap, iluju.kiringa\}@uottawa.ca} 
	}

\maketitle

\begin{abstract}
	Deep neural networks are known to be vulnerable to adversarial perturbations—small, carefully crafted inputs that lead to incorrect predictions. In this paper, we propose \textit{DeepDefense}, a novel defense framework that applies Gradient-Feature Alignment (GFA) regularization across multiple layers to suppress adversarial vulnerability. By aligning input gradients with internal feature representations, DeepDefense promotes a smoother loss landscape in tangential space, which is also known as feature space, thereby reducing the model's sensitivity to adversarial noise.
	
	We provide insights into how adversarial perturbation is decomposed into radial and tangential components and demonstrate that alignment suppresses loss variation in tangential space, where most attacks are effective. Empirically, our method achieves significant improvements in robustness across both gradient-based and optimization-based attacks. For example, on CIFAR-10, CNN models trained with DeepDefense outperform standard adversarial training by up to 15.2\% under APGD attacks and 24.7\% under FGSM attacks. Against optimization-based attacks like DeepFool and EADEN, DeepDefense requires 20-30 times higher perturbation magnitudes to cause misclassification, indicating stronger decision boundaries and flatter loss landscape. Our approach is architecture-agnostic, simple to implement, and highly effective, offering a promising direction for improving the adversarial robustness of deep learning models.
\end{abstract}
\section{Introduction}\label{sec:introduction}

Deep Neural Networks (DNNs) have achieved remarkable success in a wide range of applications, including object recognition \cite{zou2023object}, natural language processing \cite{zhou2024comprehensive}, associative memory \cite{sra2024linci, basin2023lin}, and time series regression \cite{lin2024agriculture, lin2022stacked}. Despite their high performance, DNNs are vulnerable to adversarial attacks, imperceptible input perturbations carefully crafted to mislead models into incorrect predictions \cite{nguyen2015deep}. This vulnerability poses a critical threat to the deployment of deep learning in high-stakes applications, including medical diagnosis \cite{javed2025robustness}, autonomous driving \cite{ahmed2025towards}, agricultural forecasting \cite{lin2024agriculture}, and cybersecurity \cite{al2025artificial}.

In recent years, substantial progress has been made in developing adversarial defense strategies. A prominent class is adversarial training, where models are trained on adversarial examples to enhance robustness \cite{shafahi2019adversarial}. Variants such as tradeoff-inspired Adversarial Defense via Surrogate-loss minimization (TRADES) \cite{zhang2019theoretically} and Fast Adversarial Training \cite{wongfast} offer improved trade-offs between performance and computational efficiency. Meanwhile, certified defenses aim to provide provable guarantees against attacks within bounded perturbations \cite{levine2020randomized}. Techniques like randomized smoothing \cite{cohen2019certified} and interval bound propagation (IBP) \cite{gowal2018effectiveness} have shown promise in offering formal robustness certificates.

Another line of defense involves input transformation methods, which seek to sanitize or denoise potentially adversarial inputs before classification. Examples include feature denoising networks \cite{xie2019feature} and denoising diffusion models \cite{nie2022diffusion}. In parallel, efforts to mitigate gradient obfuscation—a common pitfall in poorly designed defenses—have led to approaches like ensemble adversarial training \cite{tramer2018ensemble} and backdoor-resilient architectures \cite{liu2018fine}. More recently, researchers have begun designing defenses that target the internal representations and architecture of deep networks. These include Lipschitz-constrained models \cite{tsuzuku2018lipschitz, gouk2021regularisation}, gradient alignment regularization \cite{ross2018improving}, and robust Vision Transformers (ViTs) \cite{mao2022towards}. Finally, post-hoc detection mechanisms attempt to identify adversarial inputs after inference, using methods such as Mahalanobis distance-based anomaly detection \cite{lee2018simple} and logit-space inconsistency checks \cite{dathathri2018detecting}.

\subsection{Motivation and Contribution}
Existing defense methods, such as adversarial training \cite{shafahi2019adversarial, wongfast}, feature denoising \cite{xie2019feature}, randomization \cite{cohen2019certified}, and distillation \cite{goldblum2020adversarially}, either rely on expensive retraining or focus on isolated layers, often failing to prevent perturbation propagation through the network. To address this, we propose DeepDefense, which blocks adversarial perturbations at every layer by enforcing Gradient-Feature Alignment (GFA). This layer-wise regularization suppresses distortion propagation, offering a scalable and intrinsically robust defense.

\begin{figure}[htbp]
	\centering
	\includegraphics[width=0.69\linewidth]{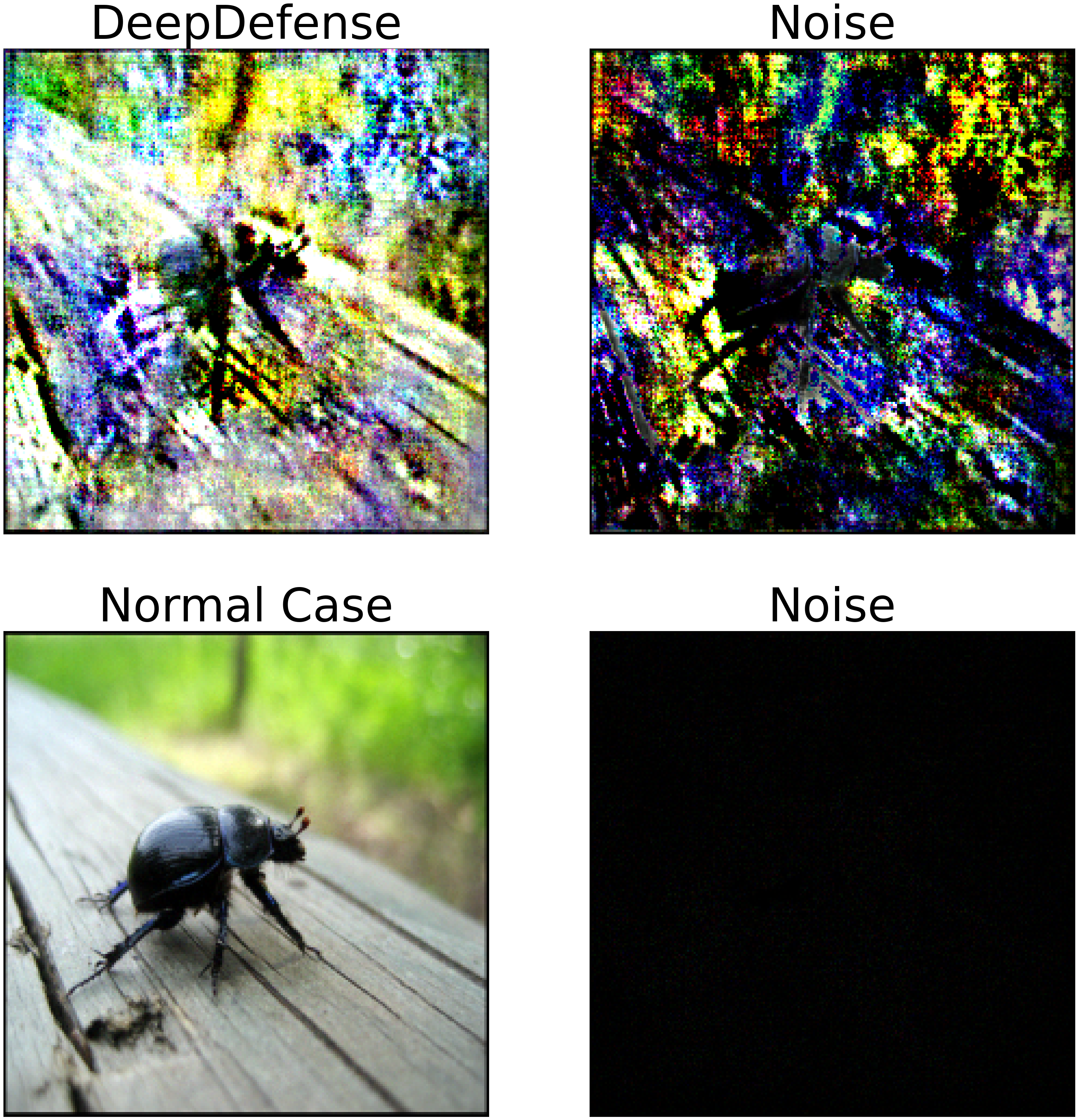}
	\caption{Visualization of adversarial perturbations generated by DeepFool. The top row shows the adversarial example (left) and corresponding perturbation (right) for a model trained with DeepDefense. The bottom row shows the same for a model trained with standard backpropagation. DeepDefense requires larger, more visible perturbations to fool the model, indicating improved robustness.}
	\label{fig:adversarial_sample}
\end{figure}

As shown in Figure \ref{fig:adversarial_sample}, for most optimization-based attack approaches, such as DeepFool, a large amount of noise (which is perceptible) along the radial space of the input sample is required to mislead the CNN trained with the DeepDefense strategy. In contrast, the model trained using standard backpropagation can be misled by adding an imperceptible amount of noise in the feature space. DeepDefense employs layer-wise GFA regularization to ensure that adversarial perturbations do not accumulate as they propagate from early convolutional layers to deeper fully connected layers, thereby increasing the model’s resistance to attacks. The main contributions of this study are as follows:

\begin{itemize}
	\item Introduce Gradient-Feature Alignment (GFA) regularization across different layers as a simple yet effective intrinsic defense that aligns input gradients with feature representations to promote a flatter loss landscape along the feature space and enhance robustness.
	\item Provide empirical analysis showing that adversarial perturbations can be decomposed into radial and tangential components relative to the input. GFA suppresses vulnerability along tangential space—where most attacks are effective—while noise in the radial space is naturally filtered by the model, resulting in improved robustness.
	\item Demonstrate enhanced robustness of GFA-trained models across a wide range of adversarial attacks, including both gradient-based and optimization-based methods, with visual and quantitative evidence confirming improved decision boundaries.
\end{itemize}

Through extensive experiments, we demonstrate that DeepDefense significantly enhances model robustness, outperforming existing adversarial defense techniques.

\subsection{Organization}
The remainder of this paper is organized as follows. Section~\ref{sec:methodology} presents the mathematical formulation of our proposed GFA regularization and its layer-wise extension, DeepDefense.  Section~\ref{sec:experiment_discussion} describes the experimental setup and evaluates the performance of DeepDefense across various adversarial attacks on CNN and MLP models.  Section~\ref{sec:conclusion} concludes the paper and outlines directions for future work. 

\section{Mathematical Formulation for DeepDefense Framework}\label{sec:methodology}

\subsection{Gradient-Feature Alignment Regularization}

\begin{definition}[Gradient Feature Alignment (GFA)]
	Gradient Feature Alignment (GFA) is a regularization principle that constrains the input gradient of the loss function to lie in the span of the input vector. 
	Formally, GFA maximizes the cosine similarity between $\mathbf{x}$ and $\nabla_{\mathbf{x}}\mathcal{L}$, as given in Eq.~\ref{equ:gfa}. 
	In the ideal case, $\nabla_{\mathbf{x}}\mathcal{L} = \lambda \mathbf{x}$ for some scalar $\lambda$, indicating perfect radial alignment.
\end{definition}

\begin{equation}\label{equ:gfa}
	\text{GFA}:  \cos\theta = \frac{\langle \mathbf{x}, \nabla_{\mathbf{x}} \mathcal{L} \rangle}{\|\mathbf{x}\| \cdot \|\nabla_{\mathbf{x}} \mathcal{L}\|} 
\end{equation}

\begin{definition}[Adversarial Perturbation]
	Let $f_\theta : \mathbb{R}^d \to \mathcal{Y}$ be a model parameterized by $\theta$, and let
	$\mathcal{L}(f_\theta(x), y)$ denote the loss function evaluated at input
	$x \in \mathbb{R}^d$ with label $y$.
	An \emph{adversarial perturbation} $\delta \in \mathbb{R}^d$ is a small,
	norm-bounded vector that maximizes the loss while remaining imperceptible, i.e.,
	\begin{equation}
		\delta^\star
		\;=\;
		\arg\max_{\|\delta\|_p \le \varepsilon}
		\mathcal{L}\big(f_\theta(x+\delta),\, y\big),
	\end{equation}
	where $\|\cdot\|_p$ denotes the $\ell_p$ norm (commonly $p \in \{2,\infty\}$),
	and $\varepsilon > 0$ controls the perturbation magnitude.
\end{definition}

\begin{definition}[Adversarial Robustness]
	Let $f_\theta : \mathbb{R}^d \to \mathcal{Y}$ be a neural network and
	$\mathcal{L}(f_\theta(x), y)$ a loss function.
	The network is said to be \emph{robust to adversarial perturbations of radius $\varepsilon$}
	at input $x$ if
	\begin{equation}
		\max_{\|\delta\|\le \varepsilon}
		\big(
		\mathcal{L}(f_\theta(x+\delta), y)
		-
		\mathcal{L}(f_\theta(x), y)
		\big)
		\le \tau,	
	\end{equation}
	where $\tau>0$ is a sufficiently small constant such that the resulting loss increase
	does not change the predicted label.
\end{definition}

\begin{remark}
	We focus on perturbations $\delta$ orthogonal to the input $x$, since perturbations parallel to $x$ primarily scale existing feature directions and tend to reinforce the signal rather than act
	adversarially. In contrast, orthogonal perturbations introduce feature-independent components
	that are more likely to induce misclassification. Therefore, robustness to orthogonal perturbations constitutes the primary notion of adversarial robustness considered in this work.
\end{remark}

\begin{lemma}[Forward Propagation of Adversarial Perturbations]
	\label{lem:perturbation_propagation} 
	
	Consider a feedforward neural network defined by
	\[
	h_0 = x, \
	z_\ell = W_\ell h_{\ell-1} + b_\ell, \
	h_\ell = \sigma_\ell(z_\ell), \ \ell = 1, \dots, L,
	\]
	where each activation function $\sigma_\ell$ is differentiable.
	
	Let $x' = x + \delta$ be a perturbed input, and define the induced feature perturbation at layer $\ell$ as:
	\begin{equation}
		\Delta h_\ell = h_\ell(x+\delta) - h_\ell(x).
	\end{equation}
	Then, under a first-order approximation of Taylor expansion,
	\begin{equation}
		\Delta h_\ell \;\approx\; J_\ell(x)\,\delta,
	\end{equation}
	where \[
	J_\ell(x) = \frac{\partial h_\ell}{\partial x}
	= \prod_{k=1}^{\ell} D_k W_k,
	\qquad
	D_k = \mathrm{diag}\!\left(\sigma_k'(z_k)\right).
	\]
	Consequently, the perturbation magnitude satisfies
	\[
	\|\Delta h_\ell\|
	\;\le\;
	\Big( \prod_{k=1}^{\ell} \|D_k W_k\| \Big)\,\|\delta\|.
	\]
\end{lemma}

Lemma~\ref{lem:perturbation_propagation} shows that input perturbations are progressively amplified through the network by the product of layer-wise Jacobian gains. If the operator norms $\|D_k W_k\|$ are large, even small input perturbations can grow exponentially with depth, leading to unstable internal representations. This observation motivates controlling perturbation amplification at each layer to improve robustness.

\begin{proposition}[GFA-Implied Robustness]\label{prop:gfa_implied_robustness}
	Let $f_\theta$ be a neural network with a differentiable loss function
	$\mathcal{L}(f_\theta(x), y)$ at an input $x \neq 0$.
	Assume that \emph{Gradient Feature Alignment (GFA)} holds at $x$, i.e.,
	\begin{equation}
		\frac{\langle x, \nabla_x \mathcal{L}(f_\theta(x), y) \rangle}
		{\|x\|\,\|\nabla_x \mathcal{L}(f_\theta(x), y)\|}
		= 1 .	
	\end{equation}
	Then, for any perturbation $\delta$ orthogonal to $x$,
	\begin{equation}
	\mathcal{L}(f_\theta(x+\delta), y)
	-
	\mathcal{L}(f_\theta(x), y)
	\;\to\; \min .
	\end{equation}
	That is, when GFA holds, the increase in loss induced by adversarial perturbations
	in directions orthogonal to the input is minimized, implying robustness to such perturbations.
\end{proposition}

\begin{proof}[Proof]
	By the first-order Taylor expansion,
	\begin{equation}
		\mathcal{L}(f_\theta(x+\delta), y) - \mathcal{L}(f_\theta(x), y)
		\approx
		\nabla_x \mathcal{L}(f_\theta(x), y)^\top \delta.
	\end{equation}
	The condition $\cos\theta = 1$ implies that
	$\nabla_x \mathcal{L}(f_\theta(x), y)$ is parallel to $x$.
	Hence, for any perturbation $\delta$ satisfying $\delta^\top x = 0$,
	\begin{equation}
		\nabla_x \mathcal{L}(f_\theta(x), y)^\top \delta = 0,
	\end{equation}
	so the first-order change in loss vanishes in tangential space,
	establishing the claimed robustness.
\end{proof}

\begin{figure}[htbp]
	\centering
	\includegraphics[width=0.7\linewidth]{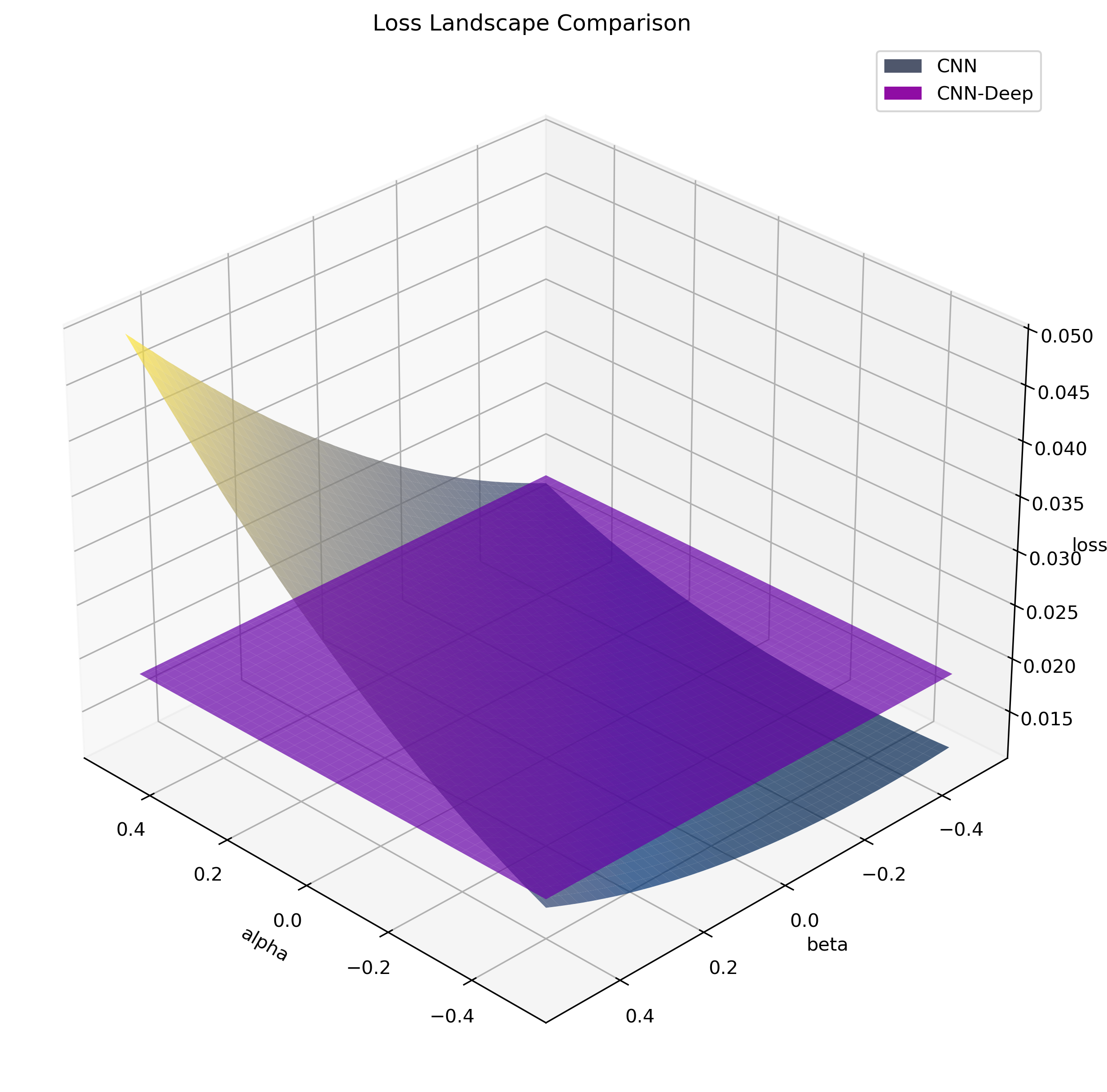}
	\caption{Loss landscapes of CNN trained with the DEEP strategy and standard backpropagation. The DEEP model exhibits a smoother and more globally flat loss surface, whereas the standard model shows relative flatness only within the radial space and undergoes significant changes under perturbations.}
	\label{fig:loss_landscape}
\end{figure}

As shown in Figure \ref{fig:loss_landscape}, two types of perturbations are generated using FGSM from models trained with and without GFA, respectively. We combine perturbations from both models in varying ratios and visualize the resulting loss landscapes for both models. It is evident that the model trained with GFA regularization exhibits a significantly flatter loss landscape compared to the standard model. This behavior arises because, in the GFA-regularized model, perturbations in the radial space are largely suppressed by activation functions or by the inherent structure of the neural network, while perturbations in the tangential space do not substantially increase the loss value, as discussed below. However, when the CNN is trained with GFA, its overall loss values tend to be higher than those of the standard CNN. As illustrated in Figure \ref{fig:loss_landscape}, the loss landscape of CNN-DEEP intersects with that of the standard CNN.

\subsection{DeepDefense: Layer-Wise Application of GFA}

Let $f_{\theta}: \mathbb{R}^d \rightarrow \mathbb{R}^K$ denote a deep neural network with $L$ layers, where $\mathbf{x} \in \mathbb{R}^d$ is the input and $\hat{\mathbf{y}} \in \mathbb{R}^K$ is the output. For a given input, the network produces a set of intermediate feature representations $F = \{F_1, F_2, \dots, F_L\},$ where $F_i$ denotes the feature tensor at the $i$-th layer.

DeepDefense improves robustness by enforcing GFA across multiple layers. Specifically, for each feature $F_i$, we compute the gradient of the task loss $\mathcal{L}_{\text{task}}$ with respect to $F_i$: $\nabla_{F_i} \mathcal{L}_{\text{task}} = \frac{\partial \mathcal{L}_{\text{task}}}{\partial F_i}.$

To suppress the propagation and amplification of adversarial perturbations, DeepDefense encourages alignment between the feature $F_i$ and its corresponding gradient $\nabla_{F_i} \mathcal{L}_{\text{task}}$. This alignment is measured using cosine similarity: $\text{GFA}_i =
\frac{
	\langle \nabla_{F_i} \mathcal{L}_{\text{task}},\; F_i \rangle
}{
	\|\nabla_{F_i} \mathcal{L}_{\text{task}}\| \, \|F_i\|
}.$

The overall training objective combines the task loss with the GFA regularization are shown in Equation \ref{equ:loss_regularization}.
\begin{equation}\label{equ:loss_regularization}
	\mathcal{L}_{\text{total}} =
	\mathcal{L}_{\text{task}}
	+
	\sum_{i=1}^{L} \beta_i \left( 1 - \text{GFA}_i \right),
\end{equation}
where $\beta_i$ controls the strength of regularization at layer $i$. The overall training procedure is summarized in Algorithm~\ref{alg:deep_defense}.

\begin{algorithm}
	\caption{DeepDefense Training}
	\label{alg:deep_defense}
	\begin{algorithmic}[1]
		\REQUIRE Training data $D = \{(x_i, y_i)\}_{i=1}^{N}$, learning rate $\eta$, model parameters $\theta$, regularization weights $\{\beta_i\}_{i=1}^{L}$
		\STATE Initialize model parameters $\theta$
		\FOR{each mini-batch $(X, Y)$ in $D$}
		\STATE $(\hat{Y}, F) \gets f_{\theta}(X)$
		\STATE $\mathcal{L}_{\text{task}} \gets \text{Loss}(\hat{Y}, Y)$
		\FOR{each feature $F_i \in F$}
		\STATE $\nabla_{F_i} \mathcal{L}_{\text{task}} \gets \frac{\partial \mathcal{L}_{\text{task}}}{\partial F_i}$
		\STATE $\text{GFA}_i \gets \frac{\nabla_{F_i} \mathcal{L}_{\text{task}} \cdot F_i}{\|\nabla_{F_i} \mathcal{L}_{\text{task}}\| \, \|F_i\|}$
		\ENDFOR
		\STATE $\mathcal{L}_{\text{total}} \gets \mathcal{L}_{\text{task}} + \sum_{i=1}^{L} \beta_i (1 - \text{GFA}_i)$
		\STATE $\theta \gets \theta - \eta \nabla_{\theta} \mathcal{L}_{\text{total}}$
		\ENDFOR
	\end{algorithmic}
\end{algorithm}

\subsection{Time Complexity Analysis}
Assuming a fully connected neural network with $L$ layers trained using a mini-batch size of $B$, let the width of layer $l$ be $d_l$. The dominant cost of one forward pass is given by $\mathcal{C} = \sum_{l=1}^{L} B\, d_{l-1} d_l.$ Since a standard backward pass has the same order of complexity, conventional backpropagation has per-batch complexity $T_{\mathrm{BP}} = \Theta(\mathcal{C}).$

For the proposed Deep Defense training, besides the standard forward pass, an additional reverse-mode differentiation is performed to compute gradients with respect to all monitored intermediate features, followed by a backward pass through the GFA term. The feature-gradient extraction requires one additional backward traversal, while the GFA computation incurs cost linear in the feature sizes, i.e., $\Theta\!\left(\sum_{i=1}^{K} |F_i|\right)$.

The final backward pass involves higher-order differentiation due to the dependence of the GFA term on $\nabla_{F_i} L$. However, this does not require explicit computation of the full Hessian matrix; instead, it relies on efficient vector-Jacobian products, whose cost remains linear in the model size. Therefore, the higher-order term satisfies $\mathcal{C}^{(2)} = \Theta(\mathcal{C})$.

Combining these components, the per-batch complexity of Deep Defense is
\[
T_{\mathrm{DD}} =
\Theta\!\left(
3\mathcal{C} + \sum_{i=1}^{K} |F_i|
\right)
=
\Theta\!\left(
\sum_{l=1}^{L} B\, d_{l-1} d_l
\right),
\]
where the feature-related term is typically lower-order. Thus, Deep Defense preserves the same asymptotic complexity as standard backpropagation, but with a larger constant factor due to additional gradient extraction and higher-order differentiation.

\section{Experiment and Discussion}\label{sec:experiment_discussion}

\subsection{Experiment Configuration}

\begin{table*}[htbp]
	\centering
	\small
	\begin{threeparttable}\setlength{\tabcolsep}{3pt} 
		\caption{GFA Regularization Values for Models Trained with Different Strategies$^{[1]}$ }
		\begin{tabular}{c|c|c|c|c|c|c}
			\hline
			Strategies & Dataset & \multicolumn{5}{c}{GFA Regularization Value} \\ \hline
			\multirow{2}{*}{FIRST} & train  & 0.9449 $\pm$ 0.0041 & 0.4207 $\pm$ 0.027 & -0.0008 $\pm$ 0.0002 & -0.0104 $\pm$ 0.0045 & -0.0473 $\pm$ 0.0896 \\ 
			& test   & 0.4734 $\pm$ 0.0112 & 0.1955 $\pm$ 0.0166 & -0.0005 $\pm$ 0.0002 & -0.0059 $\pm$ 0.0027 & -0.0257 $\pm$ 0.0545 \\ \hline
			\multirow{2}{*}{DEEP}  &  train  & 0.9262 $\pm$ 0.0039 & 0.9034 $\pm$ 0.013 & 0.9448 $\pm$ 0.0039 & -0.0077 $\pm$ 0.0025 & -0.0954 $\pm$ 0.0243 \\
			&  test   & 0.5409 $\pm$ 0.0396 & 0.5173 $\pm$ 0.038 & 0.5395 $\pm$ 0.042 & -0.005 $\pm$ 0.0022 & -0.0554 $\pm$ 0.01 \\ \hline
		\end{tabular}\label{table:gfa_all}
		\begin{tablenotes}
			\item [1] On the training dataset, the \textbf{FIRST} and \textbf{DEEP} models exhibit similar GFA values (around 0.93). However, on the testing dataset, GFA values differ significantly from those in the training dataset, suggesting that GFA may not fully generalize to unseen data in CIFAR-10 due to the presence of noisy information and the fact that the model did not learn relevant features that could be useful for classifying the testing samples. To mitigate this issue, using data augmentation to generate more training samples may help reduce this discrepancy. The GAIE loss for this model is 0.9360 $\pm$ 0.0141 on the training dataset, while on the testing dataset, it is 0.4516 $\pm$ 0.0278.
		\end{tablenotes}
	\end{threeparttable}
\end{table*}

To evaluate the effectiveness of the proposed defense mechanism, \textit{DeepDefense}, we conduct experiments on the CIFAR-10 dataset. CIFAR-10 consists of 60,000 RGB images of size $32 \times 32$, categorized into 10 classes, with 50,000 images used for training and 10,000 for testing. All input images are rescaled to the range $[-1, 1]$ prior to training and evaluation. The model is implemented using a convolutional neural network (CNN) backbone. Training is performed using the Adam optimizer with a learning rate of $1\times10^{-4}$ and a batch size of 100. The mean squared error (MSE) loss function is used for optimization. When applying the DeepDefense algorithm, the GFA values of each layer are monitored, and the training process is terminated once the model achieves optimal accuracy and the desired GFA levels.

To ensure consistent evaluation of adversarial robustness, adversarial attacks are performed only on test samples that are correctly classified by all compared models under clean conditions. This avoids bias introduced by inherently misclassified samples. Furthermore, all adversarial attacks are conducted under identical configurations across different models to ensure fair comparison.

\subsection{Case Study: Convolutional Neural Networks on CIFAR-10}

To evaluate our approach, we conduct a case study using the CIFAR-10 dataset. Six CNN models are developed, each trained with a different strategy, and evaluated under a consistent configuration against multiple adversarial attackers. For clarity, we refer to each model using the following shorthand labels:

\begin{itemize}
	\item \textbf{CNN}: Standard training using clean CIFAR-10 data.
	\item \textbf{ADV}: Adversarial training with a mix of clean and PGD-generated adversarial examples.
	\item \textbf{FIRST}: GFA regularization applied to the first convolutional layer.
	\item \textbf{DEEP}: GFA regularization applied to the first three layers.
	\item \textbf{GAIE}: GAIE regularization applied to the first convolutional layer.
	\item \textbf{DENOISE}: Feature-denoising layers inserted into the first and second convolutional layers.
\end{itemize}

For the models trained with GFA (\textbf{FIRST} and \textbf{DEEP}) as well as \textbf{GAIE}, we ensure that the corresponding regularization losses applied to the first layer are maintained at a similar scale (around 0.93) on the training dataset, as shown in Table \ref{table:gfa_all}. 

\subsubsection{Analysis of Feature Map}

\begin{figure}[htbp]
	\centering
	\includegraphics[width=1.0\linewidth]{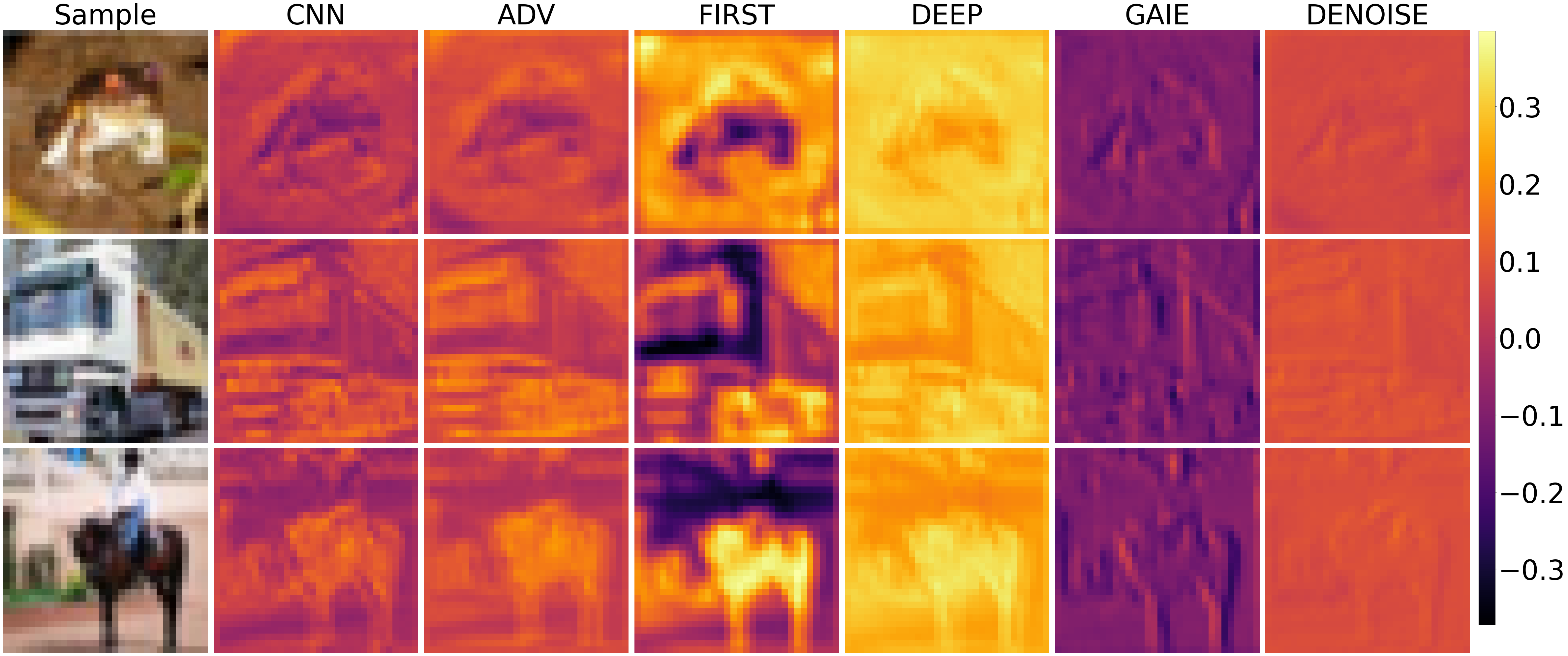}
	\caption{Feature maps from the first convolutional layer of CNNs trained with different strategies. From left to right: original input (first column), standard backpropagation, PGD adversarial training, GFA in the first layer, GFA in the first three layers, GAIE regularization, and feature denoising strategy.}
	\label{fig:feature_map}
\end{figure}

Figure \ref{fig:feature_map} illustrates the feature maps extracted from the first convolutional layer of CNNs trained with different strategies. Using the model trained by standard backpropagation (\textbf{CNN}) as a benchmark, we observe that the feature map produced by the adversarially trained model (\textbf{ADV}) is slightly shifted toward the positive activation range. The feature map of the model trained with GFA in the first layer (\textbf{FIRST}) shows increased contrast, highlighting sharper feature representations. When the alignment regularization is extended to the first three layers (\textbf{DEEP}), the feature map shifts significantly further in the positive direction, indicating enhanced and consistent activation patterns. In contrast, the model trained with GAIE regularization (\textbf{GAIE}) exhibits a notable negative shift in the feature map distribution, suggesting a different internal representation bias, likely due to its edge-focused nature. The model trained with feature denoising (\textbf{DENOISE}) produces subtle contrast feature maps, indicating suppression of both signal and noise, as mentioned in \cite{xie2019feature}.

These observations support our hypothesis that GAIE primarily captures edge features, while meaningful object representations often require more than edge information—such as color, contrast, texture and sharpness. Therefore, GFA offers a more comprehensive and robust learning signal. In fact, \textbf{GAIE} can be viewed as a special case of our GFA strategy, focusing narrowly on object edges. The enhanced robustness of the \textbf{DEEP} model over \textbf{FIRST} may be attributed to its ability to block and filter adversarial noise across multiple layers, particularly in deeper layers. This layer-wise suppression of perturbations enables the \textbf{DEEP} model to maintain more stable and meaningful feature representations under attack. Experimental results indicate that GFA applied to the first layer is most effective in mitigating adversarial perturbations, whereas its application in deeper layers offers diminished, albeit non-negligible, robustness benefits.

\subsubsection{Comparative Analysis of Model Robustness Against Gradient-Based Attacks Across Different Training Strategies}

\begin{figure}[htbp]
	\centering
	\includegraphics[width=1\linewidth]{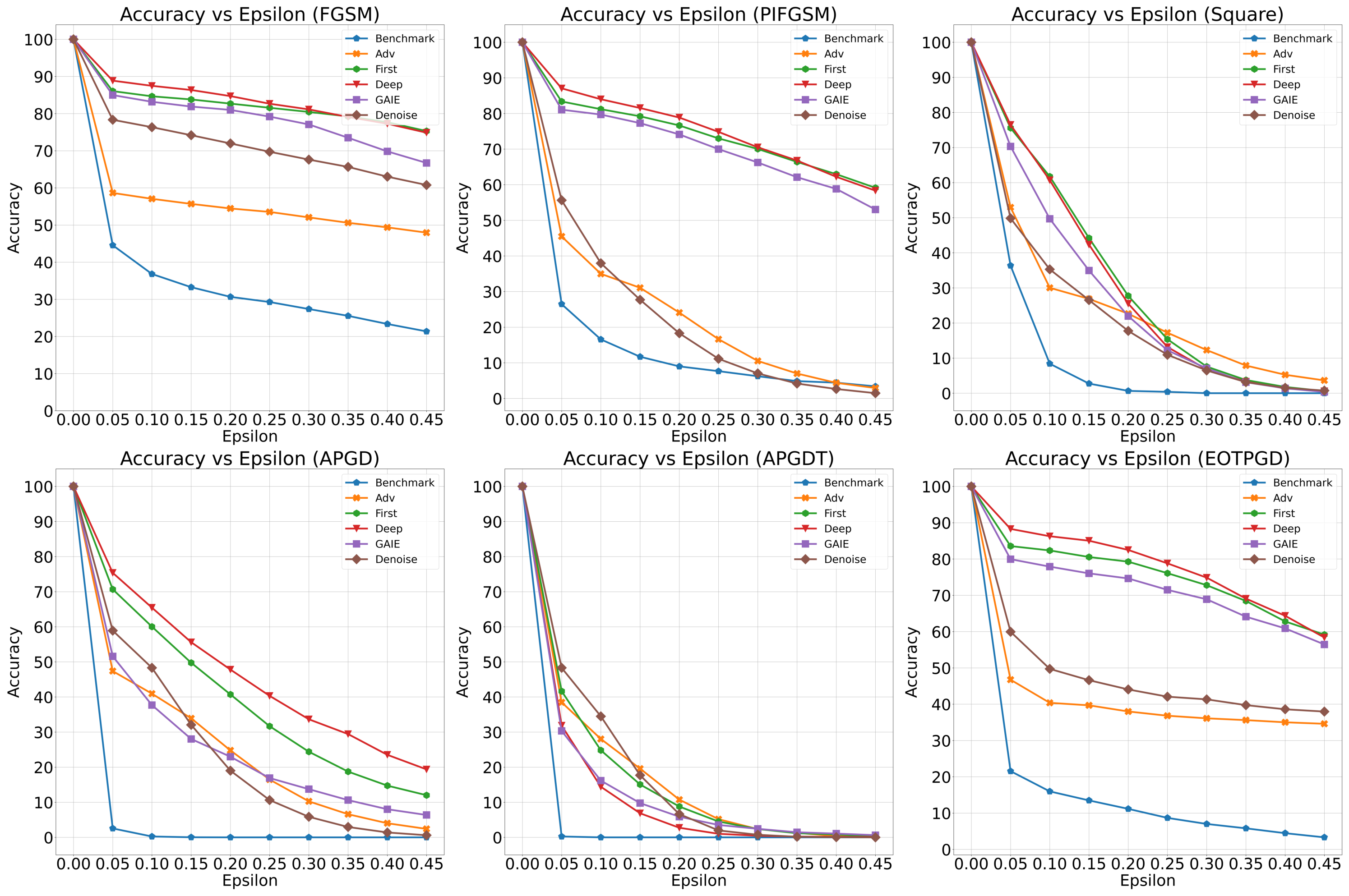}
	\caption{Evaluation of model accuracy degradation under increasing adversarial perturbation strength ($\epsilon$) for CNNs trained with different defense strategies against FGSM, PIFGSM, Square, APGD, APGDT, and EOTPGD attacks.}
	\label{fig:epislon_curve_all}
\end{figure}

\begin{table*}[htbp]
	\centering
	\scriptsize
	\begin{threeparttable}\setlength{\tabcolsep}{3pt} 
		\captionsetup{justification=centering}
		\caption{Performance of CNN Models Trained with Different Strategies \\ Against Various Adversarial Attacks on CIFAR-10}
		\begin{tabular}{c|c|c|c|c|c|c}
			\hline
			{\diagbox{Attacker$^{[1]}$}{Strategy}}  & Benchmark & PGD ADV   & First & DeepDefense$^{[2]}$  & GAIE & Denoise  \\ \cline{2-7} \hline
			FGSM & 27.42 $\pm$ 1.64 & 53.17 $\pm$ 2.45 & 77.21 $\pm$ 1.81 & \textbf{77.86} $\pm$ 2.36 & 70.26 $\pm$ 4.18 & 62.57 $\pm$ 3.48 \\ \hline
			MIFGSM & 12.48 $\pm$ 0.98 & 23.68 $\pm$ 1.5 & 74.98 $\pm$ 2.03 & \textbf{76.74} $\pm$ 3.55 & 63.56 $\pm$ 5.19 & 17.15 $\pm$ 2.83 \\ \hline
			NIFGSM & 10.87 $\pm$ 1.35 & 24.48 $\pm$ 1.25 & 67.7 $\pm$ 5.83 & \textbf{73.35} $\pm$ 3.68 & 56.22 $\pm$ 11.05 & 17.47 $\pm$ 2.75 \\ \hline
			DIFGSM & 10.03 $\pm$ 2.48 & 33.47 $\pm$ 1.56 & 54.71 $\pm$ 3.99 & \textbf{56.41} $\pm$ 4.11 & 42.37 $\pm$ 4.46 & 30.44 $\pm$ 2.64 \\ \hline
			FFGSM & 24.52 $\pm$ 1.6 & 43.56 $\pm$ 1.75 & 77.54 $\pm$ 1.64 & \textbf{79.37} $\pm$ 1.86 & 73.67 $\pm$ 2.57 & 48.53 $\pm$ 3.28 \\ \hline
			PIFGSM & 11.15 $\pm$ 1.59 & 24.8 $\pm$ 1.75 & 71.47 $\pm$ 2.63 & \textbf{73.71} $\pm$ 3.49 & 68.18 $\pm$ 4.93 & 15.34 $\pm$ 2.66 \\ \hline
			PIFGSMPP & 8.93 $\pm$ 1.69 & 21.63 $\pm$ 1.78 & 71.84 $\pm$ 2.53 & \textbf{74.31} $\pm$ 3.8 & 64.43 $\pm$ 5.2 & 13.48 $\pm$ 2.57 \\ \hline
			BIM & 12.62 $\pm$ 1.24 & 24.15 $\pm$ 1.9 & 75.07 $\pm$ 2.49 & \textbf{77.01} $\pm$ 3.48 & 64.72 $\pm$ 5.12 & 16.29 $\pm$ 2.67 \\ \hline
			EOTPGD & 12.29 $\pm$ 0.7 & 39.78 $\pm$ 1.75 & 75.34 $\pm$ 2.66 & \textbf{77.82} $\pm$ 2.6 & 68.55 $\pm$ 3.47 & 39.54 $\pm$ 3.21 \\ \hline
			PGD & 8.28 $\pm$ 1.13 & 24.62 $\pm$ 2.02 & 73.17 $\pm$ 2.57 & \textbf{74.57} $\pm$ 3.6 & 62.44 $\pm$ 4.69 & 16.55 $\pm$ 2.73 \\ \hline
			PGDL2 & 18.78 $\pm$ 0.95 & 24.96 $\pm$ 2.39 & 80.44 $\pm$ 1.46 & \textbf{84.16} $\pm$ 1.48 & 71.73 $\pm$ 3.44 & 27.63 $\pm$ 3.28 \\ \hline
			APGD & 0.12 $\pm$ 0.07 & 42.52 $\pm$ 1.2 & 55.41 $\pm$ 3.59 & \textbf{57.82} $\pm$ 5.42 & 33.79 $\pm$ 3.38 & 43.03 $\pm$ 3.29 \\ \hline
			APGDT & 0.0 $\pm$ 0.0 & 28.04 $\pm$ 1.4 & 21.53 $\pm$ 4.94 & 19.35 $\pm$ 6.93 & 15.42 $\pm$ 2.66 & \textbf{29.34} $\pm$ 2.91 \\ \hline
			Square & 8.13 $\pm$ 0.51 & 27.9 $\pm$ 4.04 & 54.34 $\pm$ 4.5 & \textbf{57.25} $\pm$ 3.27 & 45.13 $\pm$ 2.4 & 28.2 $\pm$ 4.87 \\ \hline
			AutoAttack & 86.34 $\pm$ 0.58 & 87.53 $\pm$ 1.0 & 86.36 $\pm$ 0.61 & 86.54 $\pm$ 0.7 & \textbf{87.7 $\pm$ 0.6} & 87.61 $\pm$ 0.93 \\ \hline
		\end{tabular}\label{table:adversary_attack_cnn}
		\begin{tablenotes}
			\item [1] All adversarial attacks are executed under identical configuration settings across models. Consequently, the attack intensity remains consistent for all gradient-based methods. Therefore, individual configuration details are omitted from this table for brevity.
			\item [2] To assess statistical significance, we perform paired t-tests over five independent runs for each method. Specifically, we compare DeepDefense with the strongest baseline under each attack setting. The results show that the performance improvements of DeepDefense are statistically significant ($p < 0.05$) in the majority of attack scenarios, confirming that the observed gains are not due to random variation.
		\end{tablenotes}
	\end{threeparttable}
\end{table*}

\begin{comment}
	FAB & 16.36 $\pm$ 0.82 & 51.66 $\pm$ 2.42 & 14.3 $\pm$ 4.62 & 17.49 $\pm$ 9.22 & 9.02 $\pm$ 1.19 & \textbf{64.55 $\pm$ 2.31} \\ \hline
\end{comment}

\begin{figure}[htbp]
	\centering
	\includegraphics[width=1.0\linewidth]{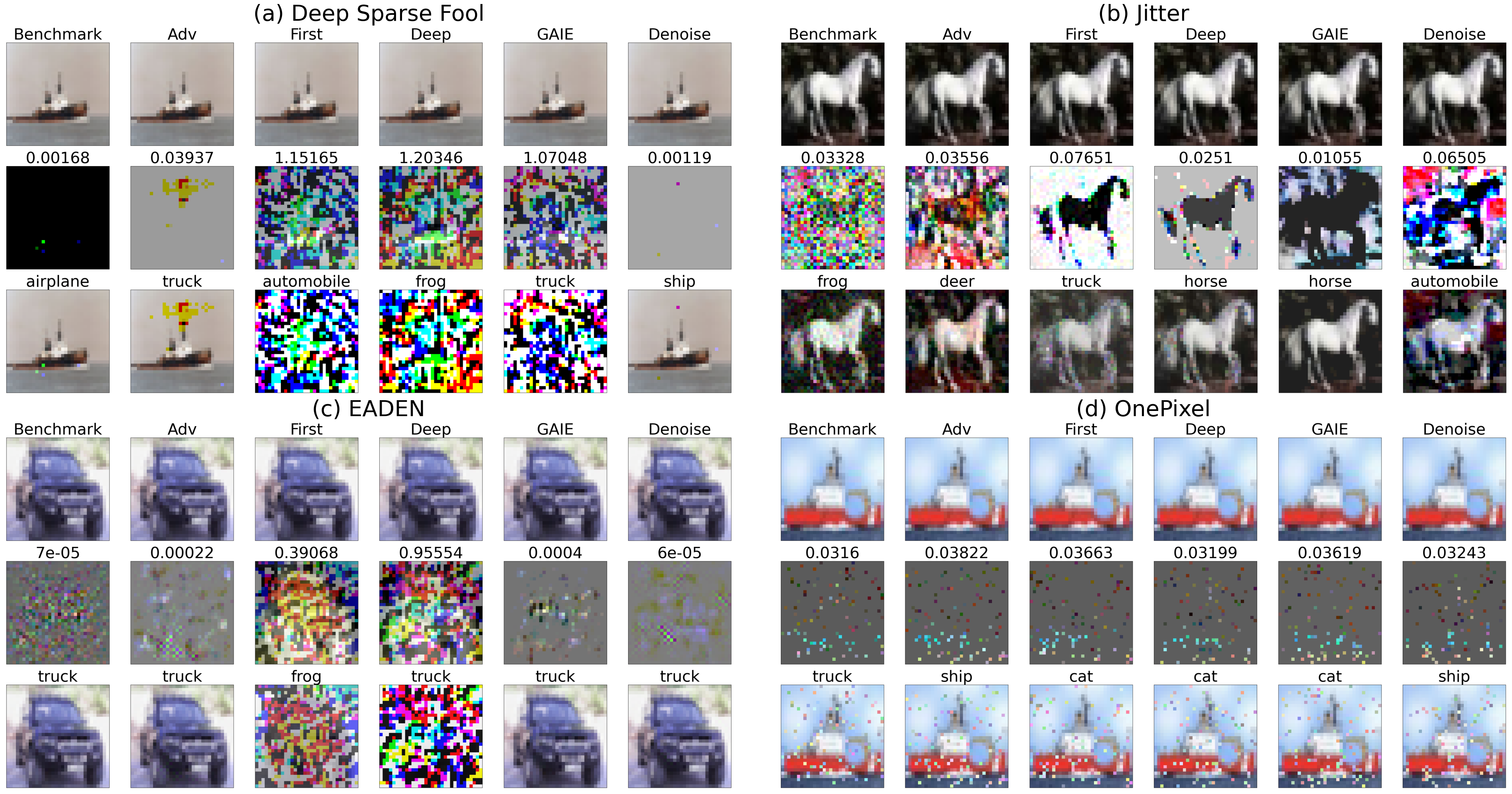}
	\caption{Comparison of the robustness of six training strategies (Benchmark, Adv, First, Deep, GAIE, Denoise) under four adversarial attacks: (a) Deep Sparse Fool, (b) Jitter, (c) EADEN, and (d) OnePixel. Each sub-image illustrates how a single sample is perturbed by one attacker and evaluated across different models. The numbers on top of the second row of each sub-figure indicate the noise intensity (measured in mean square error) applied to the perturbed samples.}
	\label{fig:cnn_optimization_based_attacker}
\end{figure}

All training strategies are repeated five times, and the average performance for each strategy is reported in Table \ref{table:adversary_attack_cnn}. We observe that the model trained with the \textbf{DEEP} strategy achieves the highest robustness across most attackers (attackers are configured with identical parameters). Exceptions occurs with the \textbf{APGDT} attacker, where the model trained using the \textbf{DENOISE} strategy demonstrates the best robustness. 

Overall, the model trained with the \textbf{DEEP} strategy consistently exhibits stronger robustness than the one trained with the \textbf{FIRST} strategy, with an improvement of approximately 1–3\% in accuracy across most attackers. The \textbf{FIRST} model, in turn, outperforms the \textbf{GAIE}, \textbf{ADV}, and \textbf{DENOISE} models, achieving roughly 5–15\% higher accuracy compared to GAIE under the majority of attack configurations. These results underscore the effectiveness of applying GFA across multiple layers in suppressing adversarial perturbations and enhancing model robustness.

Figure~\ref{fig:epislon_curve_all} complements Table~\ref{table:adversary_attack_cnn} by illustrating how model robustness evolves as the perturbation magnitude $\epsilon$ increases. The trends shown in Figure~\ref{fig:epislon_curve_all} are generally consistent with the results in Table~\ref{table:adversary_attack_cnn}. Notably, for the \textbf{APGDT} attacker, the model trained with the \textbf{DENOISE} strategy continues to outperform all other strategies when the $\epsilon$ less than 0.15. In contrast, under the \textbf{Square} attack, both the \textbf{FIRST} and \textbf{DEEP} models exhibit similar performance as $\epsilon$ increases, indicating that deeper regularization does not offer a significant advantage in this case. For the FGSM-based attackers, including FGSM and PIFGSM, models trained with \textbf{DEEP}, \textbf{FIRST}, and \textbf{GAIE} regularization consistently outperform those trained with \textbf{ADV} and \textbf{DENOISE}, especially as $\epsilon$ increases. These results further highlight the effectiveness of GFA in improving model robustness against a variety of attack strengths.

\subsubsection{Comparative Analysis of Model Robustness Against Other Attacks Across Different Training Strategies}

In Figure \ref{fig:cnn_optimization_based_attacker}, the images (a) Deep Sparse Fool, (b) Jitter, and (c) EADEN attackers illustrate adversarial examples generated by attackers that do not rely on gradient-based methods but instead use optimization-based approaches. Taking the DeepFool attacker as an example, it aims to minimize the norm of perturbation. Therefore, DeepFool formulates the attacker as a constraint optimization problem, as shown in Equation \ref{equ:perturb_input}.
\begin{equation}\label{equ:perturb_input}
	f(x + r^*) \neq f(x) \ s.t. \ \| r^* \|_p \text{ is minimized}
\end{equation}
\noindent
where \( \|\cdot\|_p \) represents the \( L_p \)-norm of the perturbation. At each iteration, the classifier is localized as a linear function around the current input \( x_t \). If we assume that the classifier is differentiable, for each class \( i \), the first-order Taylor expansion around \( x_t \) can be obtained, as shown in Equation \ref{equ:first_order_expansion}.
\begin{equation}\label{equ:first_order_expansion}
	f_i(x) \approx f_i(x_t) + \nabla f_i(x_t)^T (x - x_t)
\end{equation}
The decision boundary between the current class \( k \) (where \( k = \arg\max_i f_i(x) \) and any other class \( j \) is approximated by a hyperplane. However, in the context of model trained by GFA regularization, the direction the DeepFool searches corresponds to radial space of the input sample, which are the robustest direction naturally for most of the models. 

As a result, the generated adversarial perturbations are no longer imperceptible, as shown in Figure \ref{fig:cnn_optimization_based_attacker}. Advanced methods like C\&W, EADEN, Jitter, and JSMA refine DeepFool’s framework but fundamentally still explore radial space in the input space. However, for most of models, the radial directions is naturally be the most robust direction. Consequently, these methods tend to generate large and visible perturbations, making their adversarial examples less effective, as seen (a) Deep Sparse Fool, (b) Jitter, and (c) EADEN in Figure \ref{fig:cnn_optimization_based_attacker}. More interesting, For (b) Jitter, the generated adversarial noise is just the shape of a horse for the models trained by GFA regularizer, since the background is consistently black.

Interestingly, all models exhibit similar vulnerability under the OnePixel attack, as shown in Figure  \ref{fig:cnn_optimization_based_attacker}(d). Unlike gradient-based attacks, which exploit the local sensitivity of the model through $\nabla_x \mathcal{L}$, OnePixel explores the input space without relying on gradient information. Moreover, in high-dimensional input spaces, neural networks often exhibit non-uniform sensitivity across different input dimensions. Small, localized perturbations can therefore induce significant changes in internal feature representations, even when global perturbation norms remain small. Since such coordinate-wise vulnerabilities are shared across models with similar architectures, all methods tend to exhibit comparable performance under the OnePixel attack. This observation highlights an inherent limitation of gradient-based regularization approaches: while they effectively improve robustness against structured, gradient-driven adversaries, they probably do not provide strong protection against unstructured, gradient-free attacks. 

\section{Conclusion and Future Work}\label{sec:conclusion}
This paper proposed DeepDefense, a simple yet powerful defense strategy that applies GFA across different layers of a neural network. By aligning the gradient of the loss with the input features at each layer, our method forces adversarial perturbations to move in space that are naturally less effective. This results in a flatter and more stable loss landscape, helping models resist a wide range of attacks.

We supported our approach with both theoretical analysis and empirical experiments using CNNs. The results showed that DeepDefense outperforms standard training, adversarial training, and other regularization methods in terms of robustness. In particular, DeepDefense-trained models require stronger and more visible perturbations to be fooled, which makes them more reliable and secure.

Notably, DeepDefense is model-agnostic, lightweight to implement, and broadly compatible with standard architectures. These qualities position it as a practical and scalable solution for building more resilient deep learning systems. However, we also observe a slight drop in accuracy when DeepDefense is applied during training. We believe this degradation is primarily due to the limited size of the training dataset. In future work, we plan to address this issue by incorporating data augmentation techniques and apply DeepDefense to larger models and more complex datasets.

%\clearpage
%-----------------------------------
%  REFERENCE LIST
%----------------------------------------------------------------------------------------
\vspace{1\baselineskip}\vspace{-\parskip} % Creaters proper 4 blank line spacing.
%\footnotesize % Makes bibliography 10 pt font.
%\bibliographystyle{plain} %Can use a different style as long as it is one which uses numbered references in the text.
%\bibliography{paper}
%\LaTeX{} \cite{latex2e} is a set of macros built atop \TeX{} \cite{texbook}.
\bibliographystyle{unsrt} % We choose the "plain" reference style
\bibliography{ieee_reference}

\end{document}